\pdfoutput=1

\documentclass[11pt]{article}

\usepackage[final]{acl}

\usepackage{times}
\usepackage{latexsym}

\usepackage[T1]{fontenc}

\usepackage[utf8]{inputenc}

\usepackage{microtype}

\usepackage{inconsolata}

\usepackage{booktabs}
\usepackage{tcolorbox}
\usepackage{graphicx}
\usepackage{pifont}
\usepackage{graphicx}
\usepackage{amssymb}
\usepackage{amsmath}
\usepackage{multirow}

\newcommand{\rparagraph}[1]{\vspace{1.2mm}\noindent\textbf{#1}}
\newcommand{\rrparagraph}[1]{\vspace{1mm}\noindent\textit{#1}}

\definecolor{bluehighlight}{rgb}{0.78, 0.91, 0.95}
\definecolor{orangehighlight}{rgb}{1.0, 0.9, 0.85}

\newcommand{\colorScale}[1]{%
  \ifdim #1 pt < -80 pt\cellcolor[rgb]{1.0,0.6,0.6}
  \else\ifdim #1 pt > 50 pt\cellcolor[rgb]{1.0,0.6,0.6}
  \else\ifdim #1 pt < -40 pt\cellcolor[rgb]{1.0,0.7,0.7}%
  \else\ifdim #1 pt < -25 pt\cellcolor[rgb]{1.0,0.9,0.9}%
  \else\ifdim #1 pt < -10 pt\cellcolor[rgb]{1.0,0.97,0.97}
  \else\ifdim #1 pt < -4 pt\cellcolor[rgb]{0.95,1.0,0.95}
  \else\ifdim #1 pt < 30 pt\cellcolor[rgb]{0.85,1.0,0.85}%
  \else\cellcolor[rgb]{0.75,1.0,0.75}
  \fi\fi\fi\fi\fi\fi\fi
}

\definecolor{darkgreen}{rgb}{0.0, 0.5, 0.0}  

\title{On Generalization across Measurement Systems: LLMs Entail More Test-Time Compute for Underrepresented Cultures}

\author{
    Minh Duc Bui$^\nabla$\quad Kyung Eun Park$^\clubsuit$\quad Goran Glavaš$^\diamondsuit$ \\ 
    \textbf{Fabian David Schmidt}$^\diamondsuit$\quad \textbf{Katharina von der Wense}$^{\nabla\spadesuit}$ \\
    $^\nabla$Johannes Gutenberg University Mainz, Germany \quad $^\clubsuit$University of Mannheim, Germany \\
    $^\diamondsuit$Center For Artificial Intelligence and Data Science, University of Würzburg, Germany \\
    $^\spadesuit$University of Colorado Boulder, USA \\
    {\tt \{minhducbui, k.vonderwense\}@uni-mainz.de} \quad {\tt kyung.eun.park@uni-mannheim.de} \\
    {\tt \{fabian.schmidt, goran.glavas\}@uni-wuerzburg.de}
}

\begin{document}
\maketitle
\begin{abstract}
Measurement systems (e.g., currencies) differ across cultures, but the conversions between them are well defined so that humans can state facts using any measurement system of their choice. Being available to users from diverse cultural backgrounds, large language models (LLMs) should also be able to provide accurate information irrespective of the measurement system at hand. Using newly compiled datasets we test if this is the case for seven open-source LLMs, addressing three key research questions: 
(RQ1) What is the default system used by LLMs for each type of measurement? (RQ2) Do LLMs' answers and their accuracy vary across different measurement systems? (RQ3) Can LLMs mitigate potential challenges w.r.t. underrepresented systems via reasoning? 
Our findings show that LLMs default to the measurement system predominantly used in the data. Additionally, we observe considerable instability and variance in performance across different measurement systems. While this instability can in part be mitigated by employing reasoning methods such as chain-of-thought (CoT), this implies longer responses and thereby significantly increases test-time compute (and inference costs), marginalizing users from cultural backgrounds that use underrepresented measurement systems.

\end{abstract}

\section{Introduction}

\begin{figure}
    \centering
    \includegraphics[width=1.0\linewidth]{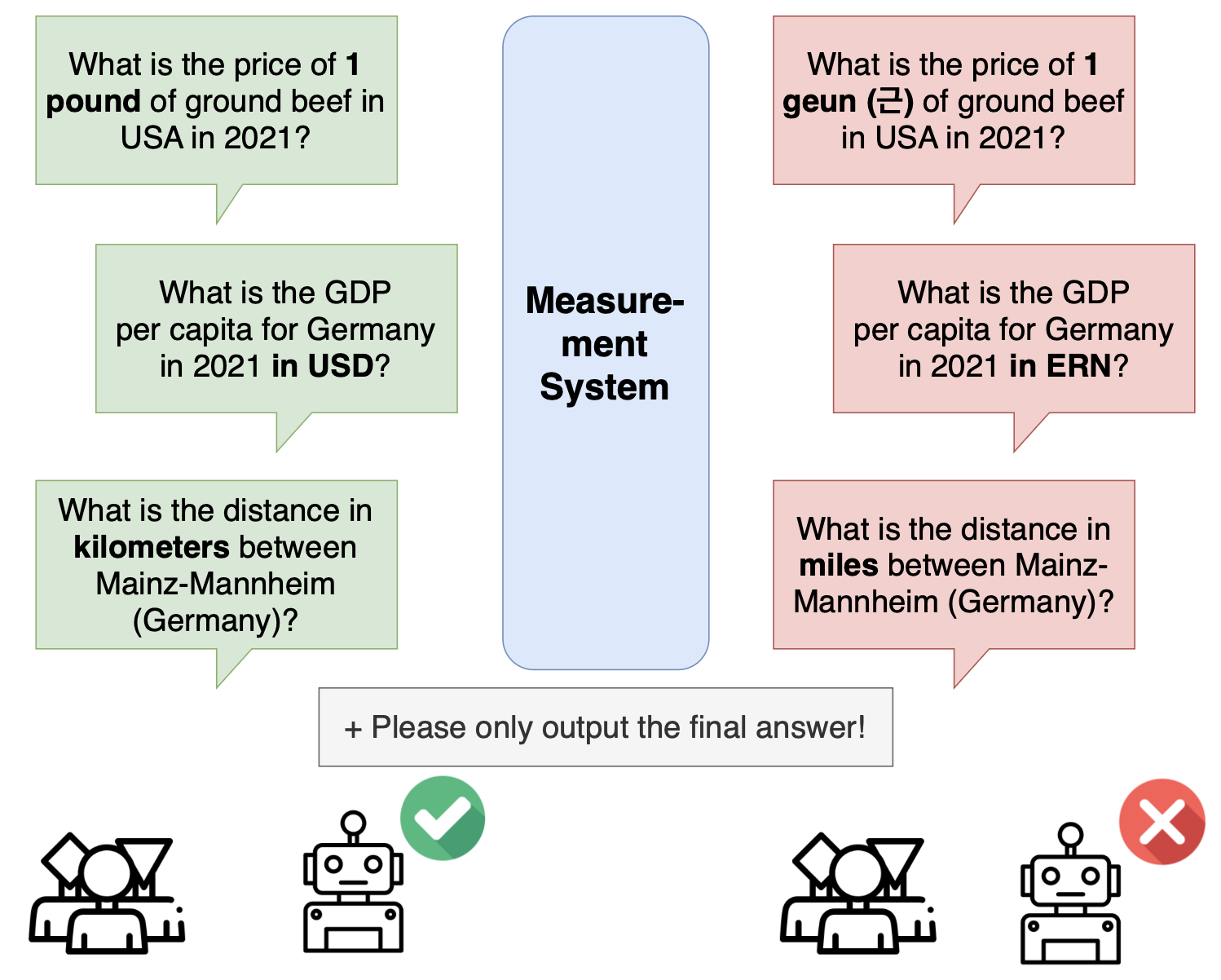}
    \caption{We test whether LLMs can translate facts across different measurement systems and show that LLMs often struggle to provide accurate information for cultural contexts underrepresented in training data.}
    \vspace{-0.4cm}
    \label{fig:intro}
\end{figure}

The user base of large language models (LLMs) expands to people from a wide range of cultural backgrounds, with each person bringing distinct perspectives to their interactions with these ubiquitous systems. Intuitively, users expect that data can be \textit{correctly} presented in ways that align with their cultural context, including measurement systems that they are used to. For instance, individuals accustomed to the metric system might find it challenging to interpret imperial units.  

Prior work shows that LLMs encode factual information, including plethora of measurement values, within their parameters \cite[\textit{inter alia}]{petroni-etal-2019-language,  geva-etal-2023-dissecting, allenzhu2024physicslanguagemodels31}. However, while an LLM may accurately present facts within one cultural context (presumably the one that dominated their respective training corpora), an inclusive system should be able to do so in \textit{all} contexts. By failing to do so, LLMs would disadvantage users from cultural backgrounds that are less represented in their training data. This raises a critical, yet underinvestigated question: can LLMs effectively generalize across socio-cultural contexts, offering same levels of factual correctness irrespective of the concrete culturally-bound factors such as measurement systems? Or do they exhibit biases that hinder socio-cultural adaptability, preventing them from equitably providing accurate information to all users?

In this work, we demonstrate that LLMs are less reliable when the users' measurement system differs from the dominant system of the LLM. We find that LLMs typically default to the measurement that fits the cultural context of the data, e.g., reporting fiscal values in USD as well as length and weight in the metric or imperial system, depending on the context of the data. For instance LLMs report food prices of US goods in USD normalized per pounds. The main issue arises when users naively prompt the models with systems that differ from the dominant one, as LLMs often fail to successfully translate values from one measurement into another (e.g., monetary amounts from USD to SDG). While advanced prompting techniques such as chain-of-thought reasoning largely mitigate these conversion errors (e.g., from 72\% to almost no difference for currency for the largest LLMs in our evaluation), we need to note that this improvement comes at a substantial cost. Under CoT, models produce much longer, more verbose responses as they triangulate the conversion process step-by-step.  For example, obtaining the correct value for the GDP of a country in Eritrean Nakfa can increase the inference cost by up to 300\%\footnote{Our estimates are based on API pricing from \url{https://groq.com/pricing/} for Llama 3.3 70B.} compared to requesting the same value in US Dollars. 
Such issues have severe implications for users from marginalized cultural backgrounds, who are meant to benefit the most from equitable access to language technology. When LLMs misrepresent or inaccurately convert critical measurements (e.g., wrongly measuring monetary value of consumer goods), they alienate users by failing to align with their cultural context. Moreover, as these users frequently come from economically disadvantaged regions, they are disproportionately affected by the increased inference costs associated with extended model responses.

We address the following research questions in this work (together with respective contributions). 

\rparagraph{(RQ1) \textit{What measurement do LLMs default to?}}  

We curate datasets that span diverse cultural contexts, including fiscal data, food prices, and city distances across different countries and regions.  
We find that LLMs default to the measurement that fits the cultural context of the data. When queried on distances between German and US cities, they provide estimates in the metric and imperial units, respectively. Likewise, LLMs express fiscal data in USD, the common convention for financial data.

\rparagraph{(RQ2) \textit{Do LLMs' answers and their accuracy vary across measurement systems?}}

We test whether varying measurement systems affects factual accuracy of answers provided by LLMs (as illustrated in Figure \ref{fig:intro}).
We find that LLMs exhibit significant variability in behavior depending on the measurement system embedded in the prompt. \textbf{Models perform best when the query aligns with the measurement system that fits the cultural context of the data (e.g., provided by a location or entity mentions)}. For instance, models show a strong decline in accuracy when asked about food prices in pounds for a country that uses kilograms.

\rparagraph{(RQ3) \textit{Can LLMs mitigate these challenges with reasoning?}}

We find that LLMs often struggle to generate accurate facts across different measurement systems. Nevertheless, very large LLMs (70B parameters) can successfully perform sequential single-hop inferences, starting with the default measurement and then converting the value to the target unit. This exemplifies that LLMs necessitate additional guidance to reliably triangulate to the correct value in the target measurement. Explicitly prompting LLMs to reason across measurement systems similarly stabilizes performance. Yet, such reasoning remains anchored in the default measurement, increasing inference costs for queries reflecting non-default cultural contexts.

\begin{table*}[t]
    \centering
    \small
    \renewcommand{\arraystretch}{0.9}
    \begin{tabular}{l|l|l|l}
          & \textbf{Fiscal Data} & \textbf{City Distances} & \textbf{Food Prices per Weight} \\ \midrule
         \textit{Measurement System} & \textit{Currency} & \textit{Length Systems} & \textit{Weight Systems}\\ 
         \quad Units & \quad USD, EUR, VND, & \quad Kilometer (Metric),  & \quad Kilogram (Metric), \\
         & \quad KRW, CNY, JPY, ... & \quad Mile (Imperial),  & \quad Pound (Imperial) \\
         & \quad  & \quad Cape Foot (South African),  & \quad Jin (Chinese),  \\
         & \quad & \quad Wa (Thai),  & \quad Geun (Korean), \\
         & \quad & \quad Li (Chinese),  & \quad Catty/Kati (Malaysian), \\
         & \quad & \quad  ... & \quad ... \\
         \quad Number of Systems & \quad 112 & \quad 10 & \quad 10 \\\midrule 
         \textit{Total Samples} & \textit{16576} & \textit{31200 (Kilometer),} & \textit{24130 (Kilogram),} \\ 
         \textit{(Facts $\times$ Systems)} & & \textit{7800 (Mile)} & \textit{3690 (Pound)} \\ 
         \quad Number of Facts & 148 & 3120 (Kilometer), 780 (Mile) & 2413 (Kilogram), 369 (Pound) \\ 
         \quad Example Question & \textit{What is GDP per capita for }  & \textit{What is the distance between}  & \textit{What is the price of one \textbf{Korean}}  \\ 
         &  \textit{Australia in 2021 in \textbf{JPY}?} & \textit{New York and Los Angeles in}  &   \textit{\textbf{Geun} of steak (beef) in Nigeria, }  \\
         & &\textit{\textbf{Li (Chinese)}?} & \textit{in Nigerian Naira (NGN)?} \\
         \bottomrule
    \end{tabular}
    \caption{\textbf{Measurement System Datasets:} Measurement system and units, as well as sample size by dataset type. For further information about each measurement system, see Appendix Table \ref{tab:measurement_systems}.}
    \label{tab:dataset}
    
\end{table*}

\section{Measurement Systems \& Datasets}

\subsection{Measurement Systems}
Each dataset is rooted in a specific cultural context. For instance, maps often use local units of measurement, international fiscal data such as GDP is typically reported in USD, and food prices are commonly expressed in local currencies. While previous studies have explored the geographical biases in LLMs' factual retrieval \cite{10.5555/3692070.3693479, 10.1145/3630106.3658967}, there has been limited investigation into whether these models can effectively adapt facts to diverse socio-cultural contexts. For example, can a model adapt fiscal data to a different currency or report city distances using an alternative length system? To address this gap, we focus on three measurement systems: \textbf{currency, length systems, and weight systems}. In Appendix \ref{ap:systems}, we describe each system in depth and report how often each remains in use nowadays.

\subsection{Datasets}

Table \ref{tab:dataset} provides an overview of the measurement systems employed in each dataset, including details on their total sizes and representative examples. This dataset is released under the CC BY-NC-ND 4.0 license. For complete details on usage terms and proper attribution, please refer to Appendix \ref{sec:terms-of-use}. Code and dataset are available at \url{https://github.com/MinhDucBui/MeasurementSystemBias}.

\rparagraph{Fiscal Data.}
To analyze currency systems, we collect international fiscal data focused on GDP per capita. We obtain the figures for 148 countries from the World Bank \cite{WorldBank_GDPpc} and specifically use 2021 data to align with the models’ knowledge cutoff, while at the same time minimizing recency biases \cite{10.1145/3630106.3658967}. We obtain the average monthly exchange rates for 2021 from the International Monetary Fund to convert these fiscal values into other currencies \cite{IMF_ExchangeRates}. To mitigate the impact of exchange rate fluctuations, we select the most favorable rate within that year, ensuring optimal performance. Note that fiscal data is typically reported in USD \cite{krugman1984international}.

\rparagraph{Food Prices per Weight: Kilogram System.}
We differentiate between countries that use the ``Kilogram'' (metric) system and the ``Pound'' (imperial) system. In countries that use the ``Kilogram'' system, we collect food commodity prices for items such as maize, rice, beans, and sugar reported per kilogram (e.g., price of 1 kg of rice), expressed in the respective local currency. This data stems from World Food Program via HDX \cite{WFP_Food_Prices}, covers 76 countries, and spans from 2010 to 2021.

\rparagraph{Food Prices per Weight: Pound System.}
We obtain food price data for United States from the Bureau of Labor Statistics \cite{BLS_CPI}. This dataset includes retail prices for a wide range of items, including fruits, meats, and other commodities, expressed in pound (e.g., the price of 1 pound of chicken).

\rparagraph{City Distances: Kilometer System.}
We begin by selecting four countries in which the ``Kilometer'' (metric) system is the primary measurement system: Germany, Russia, China and Japan. For each country, we randomly select 40 cities from among its 80 largest cities. We then compute the pairwise distances between all selected cities within each country, resulting in 3,120 distance measurements in total. The shortest straight-line distances between cities were calculated using the Haversine formula, with city coordinates sourced from \citet{simplemaps_world_cities}.

\rparagraph{City Distances: Mile System.}
For the United States, where the ``Mile'' (imperial) system applies, we likewise compile a list of 80 largest cities by population and calculate the shortest road distances as for the ``Kilometer'' system.

\section{Stability across Measurement Systems}

We begin our investigation of RQ1 by analyzing what measurement systems LLMs default to when prompted without any instruction on the use of measurement. Next, we investigate RQ2 by examining whether LLM behavior shifts when instructed to use a measurement system different from its default and whether this affects performance.

\subsection{Experimental Setup}

\rparagraph{Models.} We test instruct-finetuned LLMs from different model families and of varying sizes. Specifically, we probe Qwen2.5 (72B, 7B) \cite{qwen2025qwen25technicalreport}, Llama 3.3 (72B), Llama 3.1 (72B, 8B) \cite{grattafiori2024llama3herdmodels} and Aya Expanse (32B, 8B) \cite{dang2024ayaexpansecombiningresearch}. We group the models by their size into `Large' (>=70B) and `Small' (<70B). For more information, see Appendix \ref{ap:models}.

\rparagraph{Prompts.} To first isolate the effect of different measurement systems, we use English prompts in our experiments, ruling out the language of the prompt as a confounding factor. In \S\ref{sec:performance}, we also explore multilingual prompts. For robustness, we generate three prompt variants for each task. In each variant, the model is instructed to adhere to a specific measurement system, as illustrated by the following example:
\vspace{-0.5em}
\begin{quote}
\textit{Estimate the GDP per capita for Germany in the year 2021, \textbf{expressed in the currency ERN}. Your answer must only include the GDP per capita [...]}
\end{quote}
\vspace{-0.5em}
We refer to Appendix \ref{ap:prompts} for further details. Additionally, we specify the original name and country of origin for certain length and weight systems.

\subsection{RQ1: Default Systems} \label{sec:default}

\begin{table}[t]
    \centering
    \renewcommand{\arraystretch}{0.95}
    \small
    \begin{tabular}{l|cc|c}
         \textbf{System} & \textbf{Large} & \textbf{Small} & \textbf{Avg.} \\ \midrule
         \multicolumn{4}{c}{\textbf{Currency (Fiscal Data)}} \\ \midrule
         Local Currency & 18\%  & 16\%  & 17\% \\ 
         \textit{Other} &  &   &  \\ 
         \quad = USD & \textbf{76\%} & \textbf{82\%}  & \textbf{80\%} \\
         \quad = Rest & \underline{6\%} & \underline{2\%}  & \underline{3\%} \\ \midrule
         
         \multicolumn{4}{c}{\textbf{Weight Systems (Food Prices)}} \\ \midrule
          \textbf{Data: Kilogram} &   &   &  \\ \hline
         Kilogram &  \textbf{100\%} &  \textbf{100\%}  & \textbf{100\%} \\ 
         Pound & 0\%  & 0\%   & 0\% \\ 
         \textit{Geun, Jin, ...} & 0\%  & 0\%   & 0\% \\ \midrule
          \textbf{Data: Pound} &   &   &  \\ \hline
         Kilogram & 0\%  & 0\%  & 0\% \\
         Pound & \textbf{100\%} &  \textbf{100\%}  & \textbf{100\%} \\
         \textit{Geun, Jin, ...} & 0\%  & 0\%  & 0\% \\ \midrule
         
         \multicolumn{4}{c}{\textbf{Length Systems (City Distances)}} \\ \midrule
          \textbf{Data: Kilometer} &   &   &  \\ \hline
         Kilometer &  \textbf{100\%} &  \textbf{100\%}  & \textbf{100\%} \\
         Mile & 0\%  & 0\%  & 0\% \\
         \textit{Cape Foot, Wa, ...} & 0\%  & 0\%  & 0\% \\ \midrule

          \textbf{Data: Mile} &   &   &  \\ \hline
         Kilometer & 37\%  & 22\%  & 27\% \\ 
         Mile & \textbf{63\%} &  \textbf{78\%}  & \textbf{73\%} \\ 
         \textit{Cape Foot, Wa, ...} & 0\%  & 0\%  & 0\% \\ \bottomrule
    \end{tabular}
    \caption{\textbf{LLMs' Default Measurement Systems:} We report the total percentage of measurement systems that LLMs default to when given the freedom to choose.}
    \label{tab:default_factors}
    \vspace{-0.5em}
\end{table}

We first analyze which measurement system the model defaults to. To that end, we let the model generate its response without the instruction to output the fact in a specific measurement.

\begin{figure*}[t]
    \centering
    \includegraphics[width=1.0\linewidth]{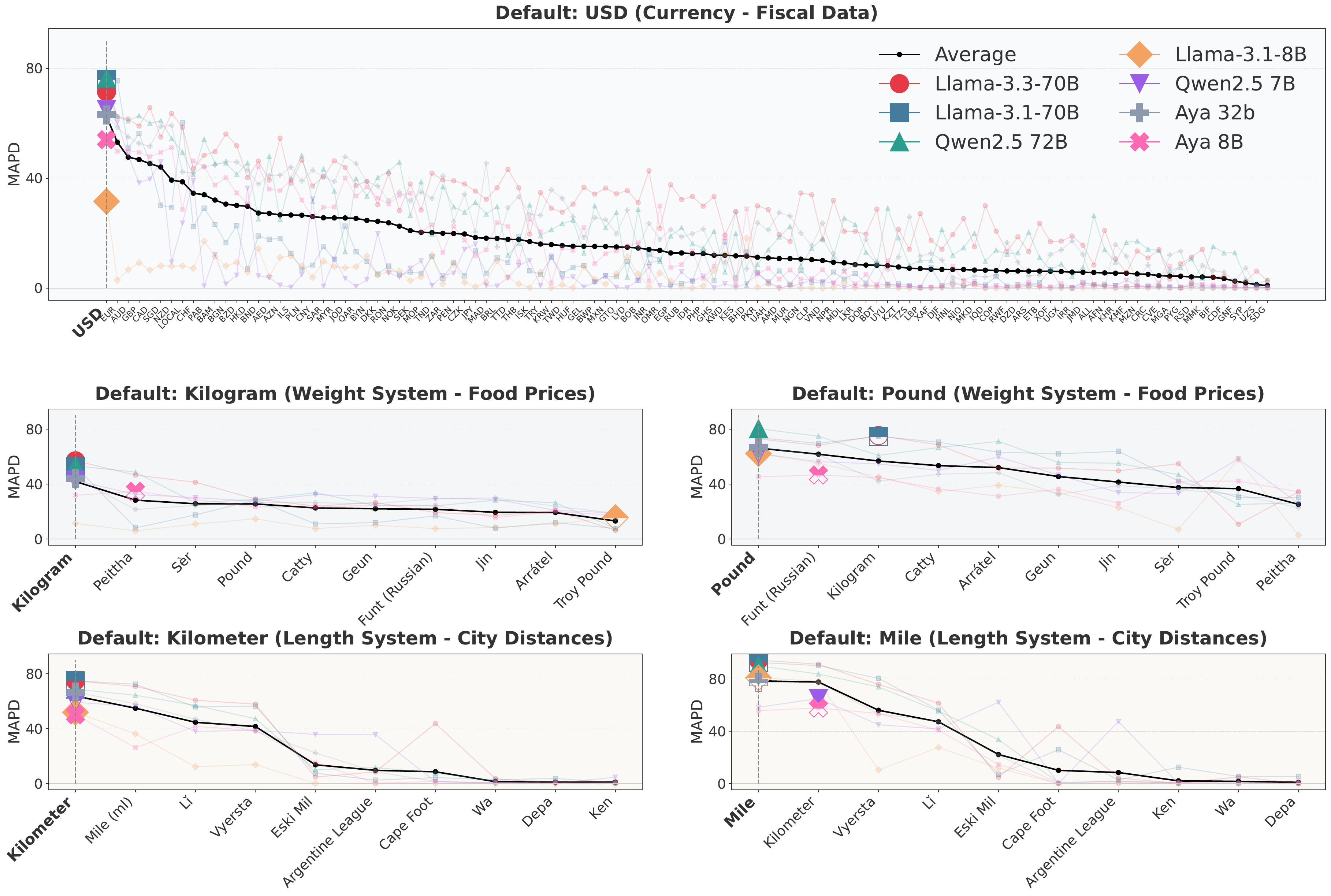}
    \caption{\textbf{Disparity and Performance of LLMs Across Different Measurement Systems:} We report the performance in terms of MAPD (cf. \S\ref{sec:performance}). The best-performing system (higher is better) for each model is highlighted with bold points. Fully filled points indicate statistically significant improvement over the second-best system, whereas half-filled points do not. We report the raw numbers and their significance in Appendix Table \ref{tab:results-appendix}.}
    \label{tab:results}
    \vspace{-0.5em}
\end{figure*}

\rparagraph{Results.} Table \ref{tab:default_factors} summarizes the findings on default systems. For fiscal data, LLMs predominantly default to using USD (80\% on average), and almost exclusively rely on either USD or local currency (17\% on average). For weight measurements, LLMs strictly employ either the imperial or metric system, in line with the data context. In the case of length data, distances in Germany, Japan, Russia and China are reported solely in the metric system, whereas, for city distances in the USA, 73\% are expressed in the imperial system.

\rparagraph{Discussion.} Our observations suggest that \textbf{LLMs tend to adopt measurement systems that mirror the conventions present in the data, often defaulting to Western standards}, such as using USD for fiscal figures. This trend seems to reflect the broader influence of the West, emerging not through any direct global mandate but through market dynamics shaped by Western forces\footnote{For example, before World War I, the pound sterling was the leading international currency; later, the dollar and the pound shared this role, and eventually the dollar dominated during the Bretton Woods era.} \cite{krugman1984international}. Interestingly, Large and Small LLMs exhibit very similar behavior, with US city distances as the only (partial) exception: Small LLMs default more consistently (+15\%) to the American cultural context, i.e., the imperial system.  

\subsection{RQ2: Disparity \& Performance across Measurement Systems} \label{sec:performance}

We evaluate the stability of LLMs' responses across measurement systems by explicitly instructing them to produce outputs in a specified system. For each dataset, the default system is the one which LLMs used most frequently in \S\ref{sec:default} (Table \ref{tab:default_factors}). We then use the performance with the default system as a baseline for the alternative systems.

\begin{figure*}[t]
    \begin{minipage}{0.65\textwidth}
        \centering
        \includegraphics[width=1.0\linewidth]{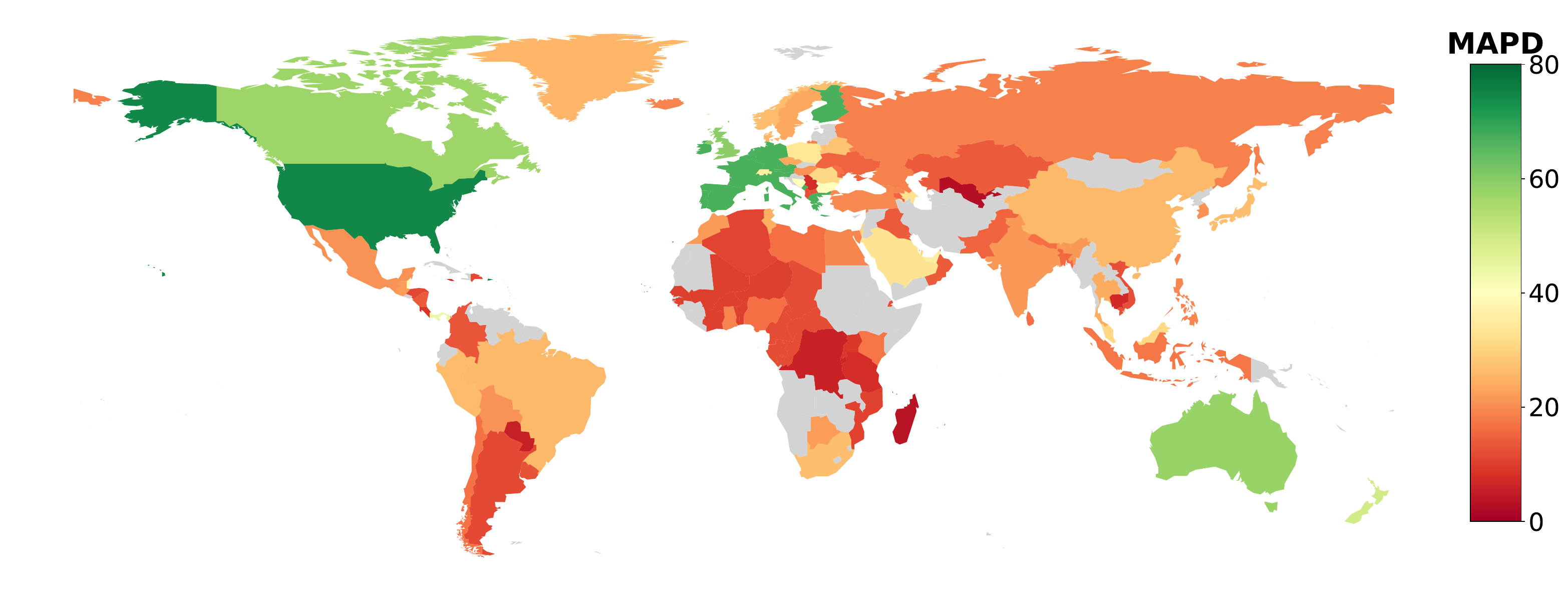}
    \end{minipage}%
    \quad
    \begin{minipage}{0.3\textwidth}
        \centering
        \small
        \begin{tabular}{c|c}
            \textbf{Income Country} & \textbf{MAPD} \\ \midrule
            High Income & \textbf{42.80\%} \\
            Upper Middle Income & 19.79\% \\
            Lower Middle Income & 14.31\% \\
            Low Income & 9.03\% \\ 
            \bottomrule
        \end{tabular}
    \end{minipage}
    \label{fig:worldmap_currency}
    \caption{\textbf{Currency Performance Mapped and Categorized by Income Groups:} We visualize the performance for all currencies—averaged across large models (over 70B parameters)—on a world map, with countries categorized by income levels.}
\end{figure*}

\rparagraph{Evaluation Metric.} We evaluate a model's performance by comparing its predictions to the ground truth. Because the magnitude of answers varies considerably, we resort to an inverse mean absolute percentage deviation (MAPD). To mitigate the impact of extreme predictions---which LLMs may occasionally produce---we cap the deviation at a maximum of 100\%. The final metric is defined as follows:

\[
APD^k_i = \min\left( \left| \frac{p(x_i, f^k)^z - y_i^z}{y_i^z} \right|, 1 \right),
\]
\[
\text{MAPD}^k = 100 \times \left(1 - \frac{1}{N} \sum_{i=1}^{N} APD^k_i \right).
\]

Here, \( p(x_i, f^k) \) is the model’s prediction for sample \( i \) using system \( k \), and \( p(x_i, f^k)^z \) represents its conversion to measurement system \( z \). The ground truth value for sample \( i \), also expressed in system \( z \), is denoted by \( y_i^z \). We note that higher values indicate better performance. An inverse MAPD of 76.0 indicates \textit{approximately} that the model’s estimates deviate from the ground truth by 24\%.

We test whether the behavior of two systems differs significantly by comparing their paired absolute percentage deviation distributions over samples. 
We employ the non-parametric Wilcoxon Signed-Rank Test \cite{wilcoxon1945individual}, which establishes whether the median difference between the paired values significantly deviates from zero. We use a high significance threshold of $p=0.001$.

\rparagraph{Results.} Table \ref{tab:results} summarizes our findings. On average across all models, the default system consistently delivers the best performance. Among 35 model–dataset combinations, it outperforms all others in 28 cases, with statistically significant gains over the second-best system in 22 of those. For example, in the fiscal dataset, using USD yields the highest performance for \textit{every} single model (average MAPD of 63\%).
Similarly, in the weight and length system dataset, models overwhelmingly yield optimal results with the default system. To illustrate, Llama 3.3 70B achieves a 57\% MAPD with the default ``Kilogram'' system, but its performance drops significantly to 29\% when using the ``Pound'' system. In the length dataset, Llama 3.3 70B achieves a 75\% MAPD on cities with the default ``Kilometer'' system and shows a 4-point performance drop when evaluated with the ``Mile'' system. Only for US cities, we observe a smaller average difference from the ``Kilometer'' system and with Qwen2.5 7B being the only model to show a statistically significant improvement when using this alternative system. While other models also perform better with certain alternative systems, their improvements are not statistically significant. Finally, we observe uneven performance drops across alternative measurement systems, as reflected in the drop of average scores.

\rparagraph{Discussion.} In sum, we find that \textbf{current LLMs are not stable across different measurement systems; LLMs yield the best performance when used with their default system and generally suffer significant and pronounced drops in performance for alternative systems}. 
Such issues have severe implications for users from marginalized cultural backgrounds, who stand to benefit the most from equitable access to language technology. When LLMs misrepresent or inaccurately convert measurements, e.g., wrongly state the value of consumer goods in local currency, they alienate users by failing to correctly adapt to their cultural context.

\rparagraph{Analysis: Currency Bias.}
We visualize currency performance of our biggest LLMs (>=70B) on a world map, associating each country with its respective currency. Additionally, we group countries by income levels provided by the World Bank \cite{fantom2016worldbank}. We immediately observe obvious disparities across income groups. LLMs achieve an average performance of 43\% with currencies from high-income countries, which is an impressive 34\% better than with currencies from low-income countries. This constitutes a clear bias that favors the richest (i.e., their currencies) and affects the poorest. As already highlighted in \S\ref{sec:performance}, using non-default systems adversely affects LLMs' performance: this analysis \textbf{reveals that this effect disproportionately disadvantages low-income countries}.

\begin{table}[t]
    \centering
    \renewcommand{\arraystretch}{0.9}
    \small
    \begin{tabular}{l|c|c|c}
        \textbf{Setting} & \textbf{Llama 3.3} & \textbf{Qwen2.5} & \textbf{Aya} \\ 
        & \textbf{70B} & \textbf{72B} & \textbf{32B}  \\ \midrule
        \multicolumn{4}{c}{\textbf{Currency (Fiscal Data)}} \\ \midrule
        \rowcolor{gray!20} \rule{0pt}{2.5ex} \textit{Default=USD} & & &  \\

        $\Delta$\textsubscript{KR} (KRW) & \cellcolor{red!20}-37.89* & \cellcolor{red!20}-27.43* & \cellcolor{red!20}-36.36* \\
        $\Delta$\textsubscript{TR} (TRY) & \cellcolor{red!20}-57.01* & \cellcolor{red!20}-42.22* & \cellcolor{red!20}-49.13* \\ \midrule

        \multicolumn{4}{c}{\textbf{Weight System (Food Prices)}} \\  \midrule       
         \rowcolor{gray!20} \rule{0pt}{2.5ex} \textit{Default=Kilogram} & & &  \\
        $\Delta$\textsubscript{KR} (Geun) & \cellcolor{red!20}-21.99* & \cellcolor{red!20}-31.12* & \cellcolor{red!20}-22.11* \\
        $\Delta$\textsubscript{ZH} (Jin) & \cellcolor{red!20}-25.39* & \cellcolor{red!20}-26.79* & \cellcolor{red!20}-11.33* \\ \midrule 
        
        \rowcolor{gray!20} \rule{0pt}{2.5ex} \textit{Default=Pound} & & &  \\
        $\Delta$\textsubscript{KR} (Geun) & -3.70 & \cellcolor{red!20}-12.54* & \cellcolor{red!20}-39.14*\\
        $\Delta$\textsubscript{ZH} (Jin) & -1.31 & \cellcolor{red!20}-20.25* & -2.29 \\ \midrule

        \multicolumn{4}{c}{\textbf{Length System (City Distances)}} \\  \midrule       
         \rowcolor{gray!20} \rule{0pt}{2.5ex} \textit{Default=Kilometer} & & &  \\
        $\Delta$\textsubscript{JA} (Ken) & \cellcolor{red!20}-80.09*  & \cellcolor{red!20}-75.71* & \cellcolor{red!20}-73.90* \\
        $\Delta$\textsubscript{ZH} (Lǐ) & \cellcolor{red!20}-20.19* & \cellcolor{red!20}-13.18* & \cellcolor{red!20}-15.03* \\ \midrule 
        
        \rowcolor{gray!20} \rule{0pt}{2.5ex} \textit{Default=Mile} & & &  \\
        $\Delta$\textsubscript{JA} (Ken) & \cellcolor{red!20}-89.68* & \cellcolor{red!20}-86.97* & \cellcolor{red!20}-84.98* \\
        
        $\Delta$\textsubscript{ZH} (Lǐ) & \cellcolor{red!20}-30.33* & \cellcolor{red!20}-35.26* & \cellcolor{red!20}-35.45* \\

    \bottomrule

    \end{tabular}
    \caption{\textbf{Aligning Language with Measurement System:} We compare performance between the default system and the system aligned with the prompt language (e.g., KRW vs. USD for Korean prompts). The language is shown as a subscript, and the aligned system in brackets. \textcolor{red}{Red} cells mark significant differences.}
    \label{tab:multilingual}
    \vspace{-0.5em}
\end{table}

\rparagraph{Analysis: Multilingual Setting.} We further examine whether language affects our findings by aligning the prompt language with the measurement system used. For instance, when using the Korean currency, we analyze the performance gap between the Korean currency and USD (default measurement system) under Korean prompt language. For each dataset, we select two languages from among Turkish (\texttt{TR}), Korean (\texttt{KR}), Chinese (\texttt{ZH}), and Japanese (\texttt{JA})---representing high- to medium-resource languages. For each language, a native speaker reviews our prompts to ensure correctness. We report the results in Table \ref{tab:multilingual}.

In nearly all cases, we observe a significant performance drop. For example, even when using the Korean currency with Korean prompt language, Llama 3.3 70B still experiences a significant decrease of 38 points. The only cases without a significant drop are Llama 3.3 70B in the weight task using the ``Pound'' system and Aya 32B when evaluating the Chinese ``Jin'' unit, also under the ``Pound'' system. We conclude that even when the language is aligned, significant drops persist in nearly all cases, highlighting the \textbf{significant instability of measurement systems independent of language in almost all cases}.

\begin{figure}[ht]
    \centering
    \includegraphics[width=1.0\linewidth]{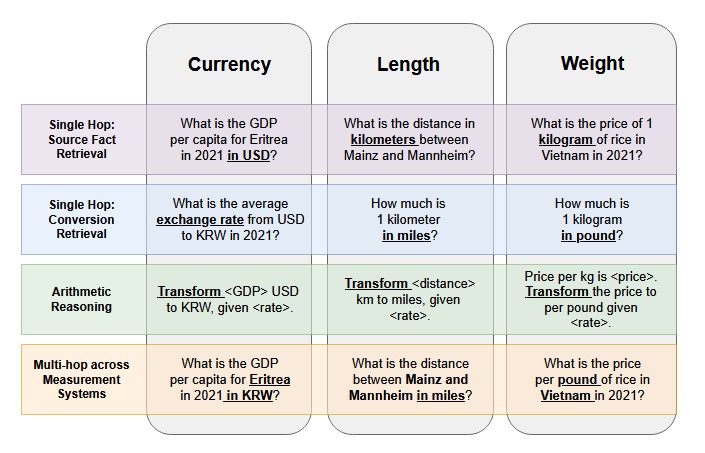}
    \caption{\textbf{Visualization of Multi-Hop Reasoning:} Answering w.r.t. an alternative measurement system requires multi-hop reasoning, consisting of the following steps: (i) retrieving the answer in the default system, (ii) retrieving the conversion rate between the default and alternative system, and (iii) combine the results of (i) and (ii) with arithmetic reasoning. }
    \label{fig:multihop}
\end{figure}

\section{RQ3: Multi-Hop Reasoning}

We first explain how answering questions w.r.t. alternative measurement systems represents a multi-hop reasoning problem and then explore whether the LLMs' performance changes with (i) a sequence of prompts that aim at individual reasoning steps and (ii) with chain-of-thought prompts, i.e., by explicitly asking the LLMs to provide the explanation for their final answer.

\rparagraph{Multi-Hop Questions.}
We adopt the definition and notation from \citet{mavi2024multihopquestionanswering}: A multi-hop question (MHQ) is a question that cannot be answered using a single document or passage (context); instead, it requires combining information from multiple sources. For example, a question like "What is the capital of the country whose currency has the highest exchange rate against the US dollar?" necessitates gathering and synthesizing information from several contexts.

Formally, let $\mathbb{C}$ denote the set of all contexts, $\mathcal{Q}$ the set of all questions, and $\mathcal{A}$ the set of all possible answers. For a question $q \in \mathcal{Q}$ and a subset of related contexts $C \subseteq \mathbb{C}$, the task is to approximate a function 
\[
f: \mathcal{Q} \times 2^{\mathbb{C}} \to \mathcal{A} \cup \{\Phi\},
\]
such that
\[
f(q, C) = 
\small{
\begin{cases} 
a \in \mathcal{A} & \text{if there exists a subset } \\
& P_q = \{p_1, \ldots, p_k\} \subseteq C, \\
&\text{with } k > 1, \text{such}\\ 
& \text{that } P_q \models (a \text{ answers } q), \\
\Phi & \text{otherwise.}
\end{cases}
}
\]
where $\models$ denotes entailment and $\Phi$ is returned when $q$ is unanswerable using $C$. The requirement that $k > 1$ ensures that the question is not solvable in a single hop (i.e., with $k = 1$, the task reduces to traditional QA).

When we ask questions of the form ``\textit{What is <Fact> in <System>?}'', where \textit{<Fact>} represents information inherently associated with a particular measurement system (e.g., GDP is intrinsically tied to currency) and the system is not the default one for the model, we are effectively asking the model to solve a multi-hop reasoning task
consisting of the following steps:

\rrparagraph{1. Source Fact Retrieval:} Extract the fact from its original measurement system (the source system, $S_{\text{src}}$). This constitutes a single-hop operation using one context, $p_1$.

\rrparagraph{2. Conversion Retrieval:} Obtain the information required to convert the fact from the source system $S_{\text{src}}$ to the target system $S_{\text{trg}}$. This is another single-hop operation that relies on a different context, $p_2$.

\rrparagraph{3. Arithmetic Reasoning:} Apply the conversion information to the retrieved fact. This requires arithmetic reasoning and may require one or more additional contexts $\{p_3, \ldots\}$.

In this scenario, the model must integrate multiple contexts $\{p1, p2, p3, ...\}$ to arrive at an answer, which aligns with our multi-hop definition. Figure \ref{fig:multihop} provides a visual overview of this process.

\begin{table*}[t]
\small
\centering
\resizebox{1.0\textwidth}{!}{%
\begin{tabular}{@{}lcccc!{\vrule}cccc!{\vrule}cc}
\toprule
 & \multicolumn{4}{c!\vrule}{\textbf{Small Models}} & \multicolumn{6}{c}{\textbf{Large Models}} \\ \cmidrule(l){2-11} 
 & \multicolumn{4}{c!{\vrule}}{\textbf{Avg. Performance}} & \multicolumn{4}{c!\vrule}{\textbf{Avg. Performance}} & \multicolumn{2}{c}{\textbf{Avg. Cost}} \\ 

\cmidrule(l){1-11} & \textit{Default Sys.} & \multicolumn{3}{c!{\vrule}}{\textit{Alternative Systems}} & \textit{Default Sys.} & \multicolumn{3}{c!{\vrule}}{\textit{Alternative Systems}}  & \multicolumn{2}{c}{\textit{Alternative Sys.}} \\ 

\cmidrule(l){1-2} \cmidrule(l){3-5} \cmidrule(l){6-6} \cmidrule(l){7-9} \cmidrule(l){10-11} 
& \textit{NoReas.} & \textit{$\Delta$NoReas.} & \textit{$\Delta$Seq.} & \textit{$\Delta$CoT} 
& \textit{NoReas.} & \textit{$\Delta$NoReas.} & \textit{$\Delta$Seq.} & \textit{$\Delta$CoT} 
& \textit{$\Delta$Seq.} & \textit{$\Delta$CoT} \\ 
\midrule

\multicolumn{11}{l}{\textbf{Default: USD (Currency Systems)}} \\ \midrule
& 53.43 & \colorScale{-77.45}\textbf{-77.45\%} & \colorScale{-31.32}\textbf{-31.32\%} & \colorScale{-44.15}\textbf{-44.15\%} 
& 74.67 & \colorScale{-72.45}\textbf{-72.45\%} & \colorScale{-0.12}\textbf{-0.12\%} & \colorScale{2.79}\textbf{+0.79\%} 
 & \colorScale{189.34}+189\% & \colorScale{302.12}+302\% \\ \midrule

\multicolumn{11}{l}{\textbf{Default: Kilogram (Weight Systems)}} \\ \midrule
& 32.43 & \colorScale{-33.11}\textbf{-33.11\%} & \colorScale{-17.92}\textbf{-17.92\%} & \colorScale{-13.88}\textbf{-13.88\%} 
& 54.33 & \colorScale{-59.13}\textbf{-59.13\%} &  \colorScale{-4.82}\textbf{-4.82\%} & \colorScale{-8.05}\textbf{-8.05\%} 
 & \colorScale{191.56}+191\% & \colorScale{206.90}+206\% \\ \midrule

\multicolumn{11}{l}{\textbf{Default: Pound (Weight Systems)}} \\ \midrule
& 59.08 & \colorScale{-33.97}\textbf{-33.97\%} & \colorScale{-19.10}\textbf{-19.10\%} & \colorScale{-33.31}\textbf{-33.31\%} 
& 75.61 & \colorScale{-29.25}\textbf{-29.25\%} & \colorScale{-4.72}\textbf{-4.72\%} & \colorScale{-9.39}\textbf{-9.39\%} 
 & \colorScale{199.77}+199\% & \colorScale{187.90}+187\% \\ \midrule

\multicolumn{11}{l}{\textbf{Default: Kilometer (Length Systems)}} \\ \midrule
& 57.00 & \colorScale{-63.64}\textbf{-63.64\%} & \colorScale{-31.82}\textbf{-31.82\%} & \colorScale{-41.74}\textbf{-41.74\%} 
& 72.84 & \colorScale{-57.93}\textbf{-57.93\%} & \colorScale{-7.88}\textbf{-7.88\%} & \colorScale{-3.98}\textbf{-3.98\%} 
 & \colorScale{180.44}+180\% & \colorScale{210.00}+210\% \\ \midrule

\multicolumn{11}{l}{\textbf{Default: Mile (Length Systems)}} \\ \midrule
& 68.13 & \colorScale{-74.36}\textbf{-74.36\%} & \colorScale{-34.55}\textbf{-34.55\%} & \colorScale{-49.95}\textbf{-49.95\%} 
& 92.46 & \colorScale{-73.13}\textbf{-73.13\%} & \colorScale{-7.18}\textbf{-7.18\%} & \colorScale{-9.70}\textbf{-7.50\%} 
 & \colorScale{197.62}+197\% & \colorScale{185.15}+185\% \\

\bottomrule
\end{tabular}
}
\caption{\textbf{Performance and Cost Increases across Reasoning Strategies:} \textit{Performance:} We report the percentage change relative to the default system (\textit{Default Sys.}) without reasoning (\textit{NoReas.}, cf. \S\ref{sec:performance}). We highlight cells in red when the performance drop exceeds 10\%. \textit{Cost Increase:} We report the additional cost incurred compared to using no reasoning.}
\label{tab:latent-composability}
\end{table*}

\rparagraph{Reasoning Chains.}
Recognizing measurement system questions as multi-hop reasoning problems, we next investigate two strategies that push LLMs to reason: (1) \textit{Sequential Single Hops} explicitly breaks down the problem for the LLM into individual reasoning steps and (2) \textit{Chain-of-Thought (CoT)} prompts implicitly nudge the LLMs to reason by requiring an explanation for the answer. 

\rrparagraph{Sequential Single Hops (Seq).} We first ask the LLMs to solve individual hops sequentially. To this end, we structure single-hop queries as illustrated in Figure \ref{fig:multihop}. We feed the hops to the LLM in sequence, integrating the answer from the previous step into the next prompt: e.g., after obtaining the answer to \textit{``What is the exchange rate from USD to VND in 2021?''} we incorporate it into the subsequent query: \textit{``Convert \texttt{<number>} USD to VND.''}

\rrparagraph{Chain-of-Thought (CoT).} We explicitly prompt the LLM to detail its intermediate steps leading to the final answer. Specifically, we prompt the model with: \textit{``What is the GDP of Germany in ERN? Explain first. Be short.''}

\begin{figure*}[t]
    \centering
    \includegraphics[width=1.0\linewidth]{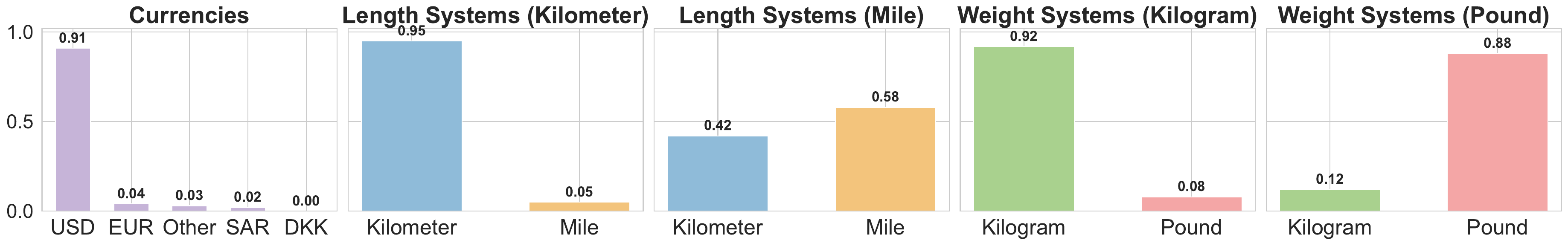}

    \caption{\textbf{Frequency of Measurement Systems as Starting Points in CoT Reasoning.} We count how often each measurement system is chosen as the starting point for the model's CoT reasoning and report the proportion. We average this proportion across all models.
    }
    \label{fig:anchor_factor}
    
\end{figure*}

\rparagraph{Evaluation.} We report the change in MAPD relative to the default measurement system prompted without any reasoning (as discussed in Section~\ref{sec:performance}), as shown in Table~\ref{tab:latent-composability}. Additionally, we examine the cost associated with these reasoning methods. For this purpose, we refer to the pricing provided by an LLM API provider\footnote{\url{https://groq.com/pricing/}} for the LLaMA 3.3 70B family and report the cost increase relative to the no-reasoning baseline  (right part of Table~\ref{tab:latent-composability}). We further examine which measurement systems, in addition to the targeted system, are referenced in the CoT reasoning (see Figure \ref{fig:anchor_factor}).

\rparagraph{Results.} Small LLMs experience a significant performance drop and show considerable instability, often failing to reason effectively, whereas Large models maintain stability across different measurement systems, with only slight performance deviations from the baseline. However, the cost of reasoning with Large models is considerably higher compared to the no-reasoning baseline. Sequential incur a cost up to 199\% of the cost without reasoning, whereas CoT on average increases the test-time compute cost to 300\% for currency conversions and around 200\% for weight and length system conversions.

\rparagraph{Discussion.} Prompting the model, explicitly (Seq) or implicitly (CoT) to reason does not benefit Small models. For Large models, reasoning does stabilize the performance to the level of the default measurement system without reasoning. This improvement comes at a significant cost: using reasoning increases test-time compute and, in turn, expenses by 180\% to 300\%, depending on the dimension of cultural context. While reasoning is an effective solution, its \textbf{increased cost disproportionately affects users who rely on alternative measurement systems rather than the LLM's default}. This discrepancy arises as models triangulate the conversion process step-by-step from the default measurement system. When faced with non-default measurement systems, \textbf{LLMs engage in additional reasoning steps to convert the result from the default system} (e.g., LLMs start from USD for currency conversion in 91\% of the cases; see Figure \ref{fig:anchor_factor}), which demands additional computation. Consequently, users with cultural contexts that are misaligned with the LLMs' default (Anglo-centric context) face higher costs---this is yet another obstacle that disproportionately hinders access to state-of-the-art language technology for the least developed parts of the world.

\section{Related Work}

\rparagraph{Biases in Factual Retrieval.} 
Recent research has examined LLMs’ factual recall \cite[\textit{inter alia}]{mallen-etal-2023-trust, 10.5555/3618408.3619049, sun-etal-2024-head, tang-etal-2023-understanding}. While these models often provide correct answers, they can hallucinate \cite{10.1145/3703155}. Studies also show that recall accuracy correlates with an entity’s popularity—with higher error rates for lesser-known subjects \cite{mallen-etal-2023-trust, sun-etal-2024-head}—and reveal geographical biases along regional and income lines \cite{10.1145/3630106.3658967, 10.5555/3692070.3693479}. 

In contrast to prior work, our study shifts attention to the systems representing facts, noting that while measurement systems may influence depiction, the underlying facts remain unchanged.  \citet{park-etal-2022-language} demonstrate that small encoder models struggle with comparisons across different measurement systems. Additionally, \citet{dinu-etal-2020-joint} explicitly incorporates measurement systems into translation models. However, unlike these studies, our work focuses on decoder-based LLMs, emphasizing factual retrieval rather than simple comparisons or translations.

\rparagraph{Cultural Biases in LLMs.} 
\citet{NEURIPS2024_be2e1b68} underscore the disparities in how LLMs are used across diverse cultural contexts. However, numerous studies have quantified and identified cultural biases in LLMs and their multimodal extensions \cite[\textit{inter alia}]{cao-etal-2024-cultural, shi-etal-2024-culturebank, NEURIPS2024_8eb88844, bui2024multi3hatemultimodalmultilingualmulticultural}. The consensus is that LLMs tend to align more closely with Western values, which limits their effectiveness in cross-cultural contexts. For instance, \citet{lee-etal-2024-exploring-cross} demonstrate that current LLMs achieve higher accuracies on hate speech labels derived from the Anglosphere. In our work, we further show that the same pattern models also favor measurement systems predominantly used in the Western world.

\rparagraph{Multi-Hop Reasoning.}
Research on multi-hop reasoning has produced numerous datasets and analyses \cite[\textit{inter alia}]{yang-etal-2018-hotpotqa, fang-etal-2020-hierarchical, chen-etal-2020-hybridqa, 10.1162/tacl_a_00475}. Recent work has shifted toward implicit multi-hop questions that require models to recall and integrate facts without explicit reasoning steps \cite{press-etal-2023-measuring, biran-etal-2024-hopping, li-etal-2024-understanding, yang-etal-2024-large-language-models}. Although LLMs can connect internal knowledge, they still struggle when key reasoning details are absent \cite{press-etal-2023-measuring, 10.5555/3666122.3669203}. Moreover, while prior work mainly examines concatenated, independent facts, our analysis shows that natural multi-hop questions from measurement systems present unique challenges.

\section{Conclusion}

In this work, we show that LLMs exhibit significant variability in performance based on the measurement system embedded in the prompt for factual retrieval. Specifically, performance declines when the model encounters non-default measurement systems. We show that reasoning (e.g., CoT) improves the model’s ability to handle alternative systems questions but it leads to an increase in test-time compute, which disproportionately affects users whose cultural contexts are misaligned with the LLMs' defaults.

\section*{Limitations}

In this study, we first prompt LLMs in English to examine how measurement systems affect factual retrieval accuracy. However, the interplay between language and measurement systems and its impact on performance remains an important avenue for future research. Users from underrepresented communities are much more likely to interact with dialogue systems in their native languages. Nonetheless, investigating how well LLMs generalize across measurement systems in English represents a `best-case' scenario, given that these models perform best in English. Consequently, our core results would be even more pronounced in other languages, where LLMs are likely to mistranslate measurements more frequently and significantly, and incur higher inference costs due to tokenization into more subwords.

Similarly, while our investigation examines a selected subset of measurement systems and units, we acknowledge the challenges of accessing comprehensive data—especially regarding systems still actively used within marginalized groups. Despite these constraints, our work successfully highlights key dynamics and interactions. We view this as a promising starting point that opens up opportunities for expanding the scope of research to include a broader array of measurement systems.

Finally, we acknowledge that there are more effective tools for retrieving facts and performing unit conversions than solely relying on LLMs, such as RAG systems or dedicated conversion tools. However, we emphasize that users are more inclined to interact with a simple LLM chatbot for these seemingly straightforward tasks rather than resorting to more complex tools.

\section*{Ethical Statement}

The sources of our data, including the World Bank, IMF, and HDX, may reflect biases inherent in data collection methodologies, economic models, or reporting structures. These biases could influence the outcomes of analyses based on the dataset. Any model trained or analyzed solely on our dataset may inherit assumptions and limitations from the data sources.

We use AI assistants, specifically o3-mini, to help edit sentences in our paper writing.

\section*{Acknowledgments}

The work of Minh Duc Bui and Katharina von der Wense was funded by the Carl Zeiss Foundation, grant number P2021-02-014 (TOPML project).


\bibliography{anthology,custom}

\appendix

\section{Dataset} 

\subsection{Measurement Systems}\label{ap:systems}
The exhaustive list of measures used in the dataset are shown in Table \ref{tab:datadisc}. 

For ease of understanding, we additionally summarize the usage context and prevalence of each system in Table \ref{tab:measurement_systems}.

\begin{table*}[t]
    \centering
    \small
    \begin{tabular}{c|c|c}
        \textbf{System} & \textbf{Usage Description} & \textbf{Usage Level} \\ \midrule

        \multicolumn{3}{c}{\textbf{Currencies}} \\ \midrule
        ALL & Currencies are used in every day life & \textit{High} \\ \midrule

        \multicolumn{3}{c}{\textbf{Weight System}} \\ \midrule
        Kilogram & Official SI base unit of mass & \textit{High} \\
        Pound & Imperial/US customary unit & \textit{High} \\
        Geun (Korea) & Korean traditional unit; still used in markets for meat. & \textit{Medium} \\
        Arrátel (Portuguese) & Historical Portuguese unit; obsolete. & \textit{Low} \\
        Funt (Russia) & Used historically in local trade and marketplaces; obsolete. & \textit{Low} \\
        Peittha & Burmese traditional unit, still used in Myanmar for local trade. & \textit{Medium} \\
        Troy pound & Specialized unit used for precious metals like gold and silver. & \textit{Medium} \\
        Jin (Mainland China) & Used in markets, and for measuring gold, silver and Chinese medicines. & \textit{Medium} \\  
        Catty/Kati (Malaysian) & Traditional weight still used in some local markets in Southeast Asia. & \textit{Low} \\
        Sèr (Pakistani) & Traditional unit used historically in the Indian subcontinent; mostly obsolete. & \textit{Low} \\ \midrule

        \multicolumn{3}{c}{\textbf{Length Systems}} \\ \midrule
        Kilometer & Used in most countries as the official system. & \textit{High} \\
        Mile & Used primarily in the US for everyday measurements. & \textit{High} \\
        Li (China) & Traditional Chinese unit of length, historically used in land measurement. & \textit{Low} \\
        Ken (Japan) & Traditional Japanese unit of length, commonly used in architecture. & \textit{Medium} \\
        Eski Mil (Turkish) & Ottoman-era unit, obsolete in modern Turkey. & \textit{Low} \\ 
        Vyersta (Russian) & Historical Russian unit; obsolete. & \textit{Low} \\
        Cape Foot (South Africa) & Colonial-era unit, obsolete in contemporary use. & \textit{Low} \\
        Depa (Indonesian) & Traditional unit based on armspan, now rarely used. & \textit{Low} \\
        Argentine league (legua) & Historical unit used for long distances in rural Argentina; obsolete. & \textit{Low} \\
        Wa (Thai) & Traditional Thai unit still used for land measurement. & \textit{Medium} \\ \midrule
    \end{tabular}
    \caption{Overview of diverse measurement systems categorized by type (Currency, Weight, and Length), including their usage descriptions and estimated level of usage. We indicate \textit{High} for systems that are officially used in the country, \textit{Medium} for those that are not official but still used in specific local contexts, and \textit{Low} for systems that are largely obsolete in everyday usage and are considered historical units.}
    \label{tab:measurement_systems}
\end{table*}

\subsection{Terms of Use} \label{sec:terms-of-use}
Our research is conducted in the public interest under GDPR, meeting the criteria for substantial academic research. To respect all data providers, our published work is licensed under CC BY-NC-ND 4.0. We use datasets with the following licenses:

\rparagraph{World Bank Group:} The fiscal data is provided under an Access to Information policy and distributed as CC‑BY 4.0.

\rparagraph{Exchange Rates:} Data from the IMF is available for download, extraction, and use under its open terms.

\rparagraph{Food Prices:} HDX and World Food Programme data use the Creative Commons Attribution for Intergovernmental Organisations license, while U.S. food prices from the Bureau of Labor Statistics are in the public domain.

\rparagraph{City Distances:} Data from the Simplemaps domain is included in approximate form for reproducibility, with proper attribution as required by their terms.

\section{Experimental Setup}

\subsection{Prompts} \label{ap:prompts}

We report all prompts used in this paper.

\rparagraph{Base Prompts (No Reasoning).}
We create three base prompt variants for the no reasoning part:

\begin{itemize}
    \item \texttt{Estimate the <MEASUREMENT> in year <YEAR>. Your answer must only include the <MEASUREMENT> as a decimal number without any abbreviations and explanation. Final number:}
    \item \texttt{Approximate the <MEASUREMENT> in year <YEAR>. Your answer must only include the <MEASUREMENT> as a decimal number without any abbreviations and explanation. Final number:}
    \item \texttt{Guess the <MEASUREMENT> in the year <YEAR>. Your answer must only include the <MEASUREMENT> as a decimal number without any abbreviations and explanation. Final number:}
\end{itemize}

\rparagraph{Base Prompts (CoT).}
To enable Chain-of-Thought (CoT) reasoning, we modify each base prompt. Specifically, we replace the instruction \texttt{"Your answer must only include the <MEASUREMENT> as a decimal number without any abbreviations and explanation. Final number:"} with \texttt{"Explain first and then give your final answer with 'Answer: <Your answer in decimal number>'. Be short:"}

\rparagraph{Variable Definitions.} For different measurement systems, the variable \texttt{<MEASUREMENT>} is defined as follows:

\begin{itemize}
    \item \textbf{Currency:} \quad \texttt{<MEASUREMENT>} = GDP per capita of \texttt{<COUNTRY>} in the currency \texttt{<CURRENCY>}
    \item \textbf{Weight Systems:} \quad \texttt{<MEASUREMENT>} = the price of \texttt{<FOOD>} per \texttt{<WEIGHTSYSTEM>} in \texttt{<COUNTRY>}
    \item \textbf{Length System:} \quad \texttt{<MEASUREMENT>} = distance between cities \texttt{<CITY1>} and \texttt{<CITY2>} in \texttt{<COUNTRY>} expressed in \texttt{<LENGTH SYSTEM>}
\end{itemize}

\subsection{Model Details} \label{ap:models}

We employ the Hugging Face implementation for all models, which are run on two A100 80GB GPUs. Outputs are generated with a limit of 40 new tokens per request, and each large LLM processes a dataset in approximately 1.5 hours with a batch size of 64. For CoT prompting, we increase the token limit, which extends the processing time to nearly 24 hours.

\begin{table*}[t]
    \centering
    \renewcommand{\arraystretch}{0.9}
    \small
    \begin{tabular}{c|c|cc|cc|cc|c}
          & \textbf{Llama 3.3} & \multicolumn{2}{c}{\textbf{Llama 3.1}} & \multicolumn{2}{|c|}{\textbf{Qwen2.5}} & \multicolumn{2}{|c|}{\textbf{Aya}} & \textbf{Average} \\ \midrule
          & \textbf{70B} &\textbf{70B} & \textbf{8B}& \textbf{72B} & \textbf{7B}& \textbf{32B} & \textbf{8B} &   \\ \midrule
          \multicolumn{9}{c}{\textbf{Currency (Fiscal Data)}} \\ 

 \hline \rowcolor{gray!20}\rule{0pt}{2.5ex}  \textit{Default = USD} & \textbf{\textit{71.51}*} & \textbf{\textit{75.99}} & \textbf{\textit{31.58}*} & \textbf{\textit{76.51}*} & \textbf{\textit{65.01}*} & \textbf{\textit{63.08}*} & \textbf{\textit{54.05}*} & \textbf{\textit{62.53}*} \\
EUR & 62.45 & 75.48 & 2.84 & 62.34 & 60.34 & 58.55 & 49.79 & 53.11 \\
AUD & 61.55 & 50.99 & 6.66 & 60.88 & 48.34 & 55.08 & 50.08 & 47.65 \\
GBP & 59.00 & 56.24 & 9.20 & 62.76 & 38.32 & 52.67 & 49.40 & 46.80 \\
CAD & 65.73 & 45.59 & 6.57 & 60.00 & 39.68 & 51.63 & 47.87 & 45.30 \\
SGD & 54.99 & 30.16 & 8.14 & 60.94 & 45.94 & 58.84 & 49.40 & 44.06 \\
NZD & 63.59 & 29.40 & 7.96 & 54.41 & 9.48 & 59.24 & 51.18 & 39.32 \\
LOCAL & 58.55 & 60.13 & 8.08 & 49.49 & 23.67 & 42.27 & 28.64 & 38.69 \\
 ... &  (all) &  (all) &   (all) &  (all) &  (all) &  (all) &  (all) & ... \\
UZS & 6.15 & 0.38 & 0.00 & 1.44 & 0.61 & 0.13 & 0.01 & 1.25 \\
SDG & 2.76 & 0.18 & 0.11 & 2.88 & 0.10 & 0.17 & 0.17 & 0.91 \\

         \midrule

          \multicolumn{9}{c}{\textbf{Weight System (Food Prices)}} \\ \hline
          
\rowcolor{gray!20} \rule{0pt}{2.5ex} \textit{Default=Kilogram} & \textbf{\textit{56.94}*} & \textbf{\textit{52.83}*} & \textit{11.33} & \textbf{\textit{53.22}*} & \textbf{\textit{42.67}*} & \textbf{\textit{43.99}*} & \textit{31.74} & \textbf{\textit{41.82}} \\
Peittha & 46.48 & 8.05 & 5.82 & 48.69 & 32.43 & 21.46 & \textbf{34.73} & 28.24 \\
Sèr & 41.28 & 17.49 & 10.85 & 26.31 & 30.10 & 24.44 & 28.81 & 25.61 \\
Pound & 29.04 & 28.44 & 14.68 & 28.81 & 27.56 & 26.43 & 23.40 & 25.48 \\
Catty & 23.95 & 10.90 & 7.57 & 33.71 & 32.27 & 26.40 & 23.13 & 22.56 \\
Geun & 26.52 & 11.91 & 10.04 & 24.36 & 31.05 & 25.54 & 24.58 & 22.00 \\
Funt (Russian) & 19.19 & 16.79 & 7.51 & 24.06 & 29.57 & 29.39 & 24.48 & 21.57 \\
Jin & 17.16 & 7.86 & 8.29 & 28.27 & 28.85 & 29.88 & 15.57 & 19.41 \\
Arrátel & 20.55 & 11.88 & 11.04 & 26.39 & 20.83 & 23.74 & 20.03 & 19.21 \\
\textbf{Troy Pound} & 6.21 & 7.61 & \textbf{15.56} & 7.58 & 18.97 & 18.77 & 17.09 & 13.11 \\ \hline

\rowcolor{gray!20} \rule{0pt}{2.5ex} \textit{Default=Pound} & \textit{72.85} & \textit{73.68} & \textbf{\textit{62.20}*} & \textbf{\textit{80.29}*} & \textbf{\textit{62.27}*} & \textbf{\textit{66.61}*} & \textit{45.25} & \textbf{\textit{66.16}} \\
Funt (Russian) & 68.15 & 69.61 & 55.22 & 74.82 & 56.25 & 61.25 & \textbf{46.61} & 61.70 \\
Kilogram & \textbf{75.32} & \textbf{74.74} & 45.12 & 60.82 & 54.66 & 42.11 & 44.56 & 56.76 \\
Catty & 68.54 & 70.46 & 34.57 & 66.56 & 49.91 & 47.11 & 36.49 & 53.38 \\
Arrátel & 51.63 & 63.14 & 38.99 & 71.05 & 59.68 & 48.24 & 31.09 & 51.98 \\
Geun & 51.59 & 62.01 & 33.56 & 55.71 & 46.59 & 32.37 & 36.24 & 45.44 \\
Jin & 49.62 & 63.81 & 22.85 & 55.12 & 33.75 & 38.81 & 25.99 & 41.42 \\
Sèr & 54.83 & 42.14 & 6.95 & 46.84 & 33.07 & 36.11 & 42.07 & 37.43 \\
Troy Pound & 10.70 & 30.20 & 57.83 & 25.12 & 58.28 & 32.14 & 42.14 & 36.63 \\
Peittha & 34.35 & 30.16 & 2.77 & 26.23 & 25.88 & 22.74 & 34.31 & 25.21 \\

         \midrule

        \multicolumn{9}{c}{\textbf{Length System (City Distances)}} \\ \hline

\rowcolor{gray!20} \textit{Default=Kilometer} \rule{0pt}{2.5ex} & \textbf{\textit{74.92}*} & \textbf{\textit{74.85}} & \textbf{\textit{51.95}*} & \textbf{\textit{68.74}*} & \textbf{\textit{59.26}} & \textbf{\textit{66.66}*} & \textbf{\textit{50.14}*} & \textbf{\textit{63.79}} \\
Mile & 70.81 & 72.42 & 36.25 & 64.39 & 56.25 & 58.14 & 26.27 & 54.93 \\
Lǐ & 60.75 & 55.83 & 12.30 & 56.92 & 38.05 & 46.94 & 41.76 & 44.65 \\
Vyersta & 57.88 & 56.47 & 13.88 & 47.28 & 38.80 & 38.33 & 38.55 & 41.60 \\
Eski Mil & 4.78 & 7.64 & 0.11 & 10.07 & 35.81 & 22.41 & 15.24 & 13.72 \\
Argentine League & 8.58 & 2.55 & 0.20 & 11.71 & 35.71 & 8.60 & 0.41 & 9.68 \\
Cape Foot & 43.77 & 4.41 & 0.00 & 7.79 & 1.37 & 1.61 & 1.51 & 8.64 \\
Wa & 3.65 & 2.72 & 0.10 & 2.03 & 0.42 & 0.37 & 0.79 & 1.44 \\
Depa & 0.32 & 3.77 & 0.14 & 1.71 & 0.23 & 0.50 & 0.99 & 1.10 \\
Ken & 0.27 & 0.70 & 0.07 & 0.60 & 4.72 & 0.27 & 0.67 & 1.04 \\

\hline
\rowcolor{gray!20} \rule{0pt}{2.5ex} \textit{Default=Mile} & \textbf{\textit{94.54}*} & \textbf{\textit{92.72}} & \textbf{\textit{80.96}} & \textbf{\textit{90.11}*} & \textit{58.40} & \textbf{\textit{77.25}} & \textit{55.92} & \textbf{\textit{78.56}} \\
Kilometer & 91.35 & 90.31 & 78.40 & 83.88 & \textbf{65.10*} & 77.24 & \textbf{57.84} & 77.73 \\
Vyersta & 75.83 & 80.50 & 10.43 & 73.97 & 45.05 & 53.45 & 53.66 & 56.13 \\
Lǐ & 61.57 & 56.29 & 27.66 & 55.53 & 41.82 & 47.46 & 40.84 & 47.31 \\
Eski Mil & 4.78 & 7.12 & 11.84 & 33.70 & 62.27 & 21.59 & 14.74 & 22.29 \\
Cape Foot & 43.72 & 25.95 & 0.01 & 0.84 & 0.12 & 0.39 & 0.10 & 10.16 \\
Argentine League & 4.24 & 2.89 & 0.34 & 1.87 & 47.48 & 2.02 & 0.61 & 8.49 \\
Ken & 0.27 & 12.56 & 0.03 & 0.26 & 0.15 & 0.76 & 0.75 & 2.11 \\
Wa & 5.15 & 5.48 & 0.19 & 0.52 & 0.17 & 0.31 & 0.27 & 1.73 \\
Depa & 0.22 & 5.46 & 0.14 & 0.26 & 0.12 & 0.40 & 0.38 & 1.00 \\

    \bottomrule

    \end{tabular}
    \caption{\textbf{Disparity and Performance of LLMs Across Different Measurement Systems:} The first row (shaded in gray) within each group represents the model's default measurement system. We report the performance in terms of MAPD (cf. \S\ref{sec:performance}). The best performance in each column (higher is better) are shown in \textbf{bold}. \textit{Disparity:} The best values are additionally marked with *, if statistically significant differences (\( p < 0.001 \)) compared to the second best system exists. Cells highlighted in \textcolor{red}{red} indicate cases where a significant difference was observed.}
    \label{tab:results-appendix}
    \vspace{-0.5em}
\end{table*}

\begin{table*}[t]
    \centering
    \tiny
    \begin{tabular}{l|l|l|l|l|l}
    \toprule
    \multicolumn{6}{c}{\textbf{Currency (Fiscal Data)}} \\ \midrule
    Currency code & Description & Conversion (to USD) & Currency code & Description & Conversion (to USD)\\ \midrule
   \textbf{AED} & United Arab Emirates Dirham & 3.6725
       &\textbf{KMF} & Comorian Franc & 418.0677243\\
    \textbf{AFN} & Afghan Afghani & 77.41418667 & \textbf{KRW} & South Korean Won & 1149.108333 
       \\
    \textbf{ALL} & Albanian Lek & 104.0547222 & \textbf{KWD} & Kuwaiti Dinar & 0.301641667
       \\
    \textbf{AMD} & Armenian Dram & 503.6105556 & \textbf{KZT} & Kazakhstani Tenge & 425.9513889
      \\
    \textbf{ARS} & Argentine Peso & 95.18694444 & \textbf{LBP} & Lebanese Pound & 1507.5 
       \\
    \textbf{AUD} & Australian Dollar & 1.337372732 & \textbf{LKR} & Sri Lankan Rupee & 201.3890667
      \\
    \textbf{AZN} & Azerbaijani Manat & 1.7
      &  \textbf{LYD} & Libyan Dinar & 4.523094444\\
    \textbf{BAM} & Bosnia-Herzegovina Convertible  & 1.6673125
      & \textbf{MDL} & Moldovan Leu & 17.66990278 \\
      &Mark&&\textbf{MAD} & Moroccan Dirham & 9.045236111 \\
    \textbf{BDT} & Bangladeshi Taka & 85.24412794
      & \textbf{MGA} & Malagasy Ariary & 3854.072778 \\
    \textbf{BGN} & Bulgarian Lev & 1.661877778
      & \textbf{MKD} & Macedonian Denar & 52.55698889 \\
    \textbf{BHD} & Bahraini Dinar & 0.376
      & \textbf{MMK} & Myanma Kyat & 1610.540741 \\
    \textbf{BIF} & Burundian Franc & 1978.428889
      & \textbf{MOP} & Macanese Pataca & 8.008919444 \\
    \textbf{BND} & Brunei Dollar & 1.345794444
      & \textbf{MUR} & Mauritian Rupee & 41.72418333 \\
    \textbf{BOB} & Bolivian Boliviano & 6.91
      & \textbf{MXN} & Mexican Peso & 20.654375 \\
    \textbf{BRL} & Brazilian Real & 5.368375
      & \textbf{MYR} & Malaysian Ringgit & 4.143145833 \\
    \textbf{BWP} & Botswanan Pula & 11.19945833
      & \textbf{MZN} & Mozambican Metical & 66.0225 \\
    \textbf{BYN} & Belarusian Ruble & 2.533175
      & \textbf{NGN} & Nigerian Naira & 406.2341667 \\
    \textbf{BZD} & Belize Dollar & 2
      & \textbf{NIO} & Nicaraguan Cordoba & 35.20105833 \\
    \textbf{CAD} & Canadian Dollar & 1.251083333
      & \textbf{NOK} & Norwegian Krone & 8.631944444 \\
    \textbf{CDF} & Congolese Franc & 1989.948356
      & \textbf{NPR} & Nepalese Rupee & 118.7686111 \\
    \textbf{CHF} & Swiss Franc & 0.9150375
      & \textbf{NZD} & New Zealand Dollar & 1.414869444 \\
    \textbf{CLP} & Chilean Peso & 779.4822222
      & \textbf{OMR} & Omani Rial & 0.3845 \\
    \textbf{CNY} & Chinese Yuan & 6.457772222
      & \textbf{PAB} & Panamanian Balboa & 1 \\
    \textbf{COP} & Colombian Peso & 3784.358056
      & \textbf{PEN} & Peruvian Nuevo Sol & 3.892861111 \\
    \textbf{CRC} & Costa Rican Colon & 625.7229167
      & \textbf{PHP} & Philippine Peso & 49.43672222 \\
    \textbf{CVE} & Cape Verdean Escudo & 93.765085
      & \textbf{PKR} & Pakistani Rupee & 164.1562869 \\
    \textbf{CZK} & Czech Republic Koruna & 21.67958333
      & \textbf{PLN} & Polish Zloty & 3.889841667 \\
    \textbf{DJF} & Djiboutian Franc & 177.721
      & \textbf{PYG} & Paraguayan Guarani & 6676.519722 \\
    \textbf{DKK} & Danish Krone & 6.317175
      & \textbf{QAR} & Qatari Rial & 3.64 \\
    \textbf{DOP} & Dominican Peso & 57.25499444
      & \textbf{RON} & Romanian Leu & 4.192530556 \\
    \textbf{DZD} & Algerian Dinar & 135.6026
      & \textbf{RSD} & Serbian Dinar & 99.94511944 \\
    \textbf{EGP} & Egyptian Pound & 15.65245278
      & \textbf{RUB} & Russian Ruble & 73.53540833 \\
    \textbf{ETB} & Ethiopian Birr & 44.25473611
      & \textbf{RWF} & Rwandan Franc & 991.6896025 \\
    \textbf{EUR} & Euro & 0.84978685
      & \textbf{SAR} & Saudi Riyal & 3.75 \\
    \textbf{GBP} & British Pound Sterling & 0.72827027
      & \textbf{SDG} & Sudanese Pound & 299.4778056 \\
    \textbf{GEL} & Georgian Lari & 3.25225
      & \textbf{SEK} & Swedish Krona & 8.649225 \\
    \textbf{GHS} & Ghanaian Cedi & 5.851411111
      & \textbf{SGD} & Singapore Dollar & 1.345016667 \\
    \textbf{GNF} & Guinean Franc & 9933.2195
      & \textbf{SYP} & Syrian Pound & 1256 \\
    \textbf{GTQ} & Guatemalan Quetzal & 7.7412875
      & \textbf{THB} & Thai Baht & 32.01021944 \\
    \textbf{HKD} & Hong Kong Dollar & 7.776777778
      & \textbf{TND} & Tunisian Dinar & 2.805069444 \\
    \textbf{HNL} & Honduran Lempira & 24.0325
      & \textbf{TRY} & Turkish Lira & 9.808511111 \\
    \textbf{HUF} & Hungarian Forint & 304.9938889
      & \textbf{TTD} & Trinidad and Tobago Dollar & 6.747011111 \\
    \textbf{IDR} & Indonesian Rupiah & 14333.63958
      & \textbf{TWD} & New Taiwan Dollar & 28.05294444 \\
    \textbf{ILS} & Israeli New Sheqel & 3.224777778
      & \textbf{TZS} & Tanzanian Shilling & 2296.881064 \\
    \textbf{INR} & Indian Rupee & 73.82430833
      & \textbf{UAH} & Ukrainian Hryvnia & 27.27136389 \\
    \textbf{IQD} & Iraqi Dinar & 1450
      & \textbf{UGX} & Ugandan Shilling & 3601.599952 \\
    \textbf{IRR} & Iranian Rial & 42000
      & \textbf{USD} & US Dollar & 1 \\
    \textbf{ISK} & Icelandic Krona & 125.9425
      & \textbf{UYU} & Uruguayan Peso & 43.51508333 \\
    \textbf{JMD} & Jamaican Dollar & 150.3607333
      & \textbf{UZS} & Uzbekistan Som & 10646.02833 \\
    \textbf{JOD} & Jordanian Dinar & 0.71
      & \textbf{VND} & Vietnamese Dong & 23177.22222 \\
    \textbf{JPY} & Japanese Yen & 109.6466667
      & \textbf{XAF} & CFA Franc BEAC & 557.4236325 \\
    \textbf{KES} & Kenyan Shilling & 110.2038889
      & \textbf{XOF} & CFA Franc BCEAO & 557.4236325 \\
    \textbf{KHR} & Cambodian Riel & 4097.125
      & \textbf{ZAR} & South African Rand & 14.955069 \\
\midrule
    \multicolumn{6}{c}{\textbf{Weight System}} \\ \midrule
     Measure & Description & Conversion (in KG) & Measure & Description & Conversion (in KG)\\ \midrule
          \textbf{Kilogram (kg)} &  & 1 & \textbf{Geun (Korea)} & & 0.6 \\
          \textbf{Pound (lb)} & & 0.453592 & \textbf{Troy pound} & & 0.37324 \\
          \textbf{Arrátel (Portugese)} & & 0.459 & \textbf{Funt (Russia)} & & 0.409517\\
          \textbf{catty/kati (Malaysian)} & & 0.60479  & 
          \textbf{Sèr (80 Tolä, Pakistani)} & & 0.93310 \\
          \textbf{Jin (Mainland China)} & & 0.5 &
          \textbf{Peittha (Burmamese)} & & 1.63293 \\
          \midrule
    \multicolumn{6}{c}{\textbf{Length System}} \\ \midrule
     Measure & Description & Conversion (in KM) & Measure & Description & Conversion (in KM)\\ \midrule
    \textbf{Kilometer (km)} & & 1 &
    \textbf{Mile (mi)} & & 1.60934 \\
    \textbf{Lǐ (China)} & & 0.5 &
    \textbf{Eski Mil (Turkish)} &  & 1.89435 \\
    \textbf{Vyersta (Russian)} && 1.0668 &
    \textbf{Ken (Japan)} &  & 0.001818 \\
    \textbf{Cape Foot (South Africa)} & & 0.0003148584 &
    \textbf{Depa (Indonesian)} & & 0.0016 \\
    \textbf{Argentine league (legua)} &  & 5.572 &
    \textbf{Wa (Thai)} & & 0.002 \\
         \bottomrule
    \end{tabular}
    \caption{\textbf{Measurement System Description.} Description of measures used in the dataset.}
    \label{tab:datadisc}
\end{table*}

\end{document}